\documentclass{llncs}
\usepackage{graphicx}
\usepackage{multirow}
\usepackage{booktabs}
\usepackage{hyperref}
\usepackage{color}
\usepackage{tabto}
\usepackage{makecell}
\begin{document}

\title{Application of a Hybrid Bi-LSTM-CRF model to the task of Russian Named Entity Recognition}
\titlerunning{Hybrid Bi-LSTM-CRF for Russian NER}  
%
\author{Anh L. T.\inst{1, 2}, Arkhipov M. Y.\inst{1}, Burtsev M. S.\inst{1}}
\institute{Neural Networks and Deep Learning Lab, Moscow Institute of Physics and Technology, Russia\\
\email{\{burtcev.ms, arkhipov.mu\}@mipt.ru}
\and
Faculty of Information Technology, Vietnam Maritime University, Viet Nam
\email {anhlt@vimaru.edu.vn}}

\maketitle              

\begin{abstract}
Named Entity Recognition (NER) is one of the most common tasks of the natural language processing. The purpose of NER is to find and classify tokens in text documents into predefined categories called tags, such as person names, quantity expressions, percentage expressions, names of locations, organizations, as well as expression of time, currency and others. Although there is a number of approaches have been proposed for this task in Russian language, it still has a substantial potential for the better solutions. In this work, we studied several deep neural network models starting from vanilla Bi-directional Long Short Term Memory (Bi-LSTM) then supplementing it with Conditional Random Fields (CRF) as well as highway networks and finally adding external word embeddings. All models were evaluated across three datasets: Gareev's dataset, Person-1000 and FactRuEval-2016. We found that extension of  Bi-LSTM model with CRF significantly increased the quality of predictions. Encoding input tokens with  external word embeddings reduced training time and allowed to achieve state of the art for the Russian NER task.
\end{abstract}
\section{Introduction}
There are two main approaches to address the named entity recognition (NER) problem \cite{Ms. Maithilee et al.}. The first one is based on handcrafted rules, and the other relies on statistical learning. The rule based methods are primarily focused on engineering a grammar and syntactic extraction of patterns related to the structure of language. In this case, a laborious tagging of a large amount of examples is not required. The downsides of fixed rules are poor ability to generalize and inability to learn from examples. As a result, this type of NER systems is costly to develop and maintain. Learning based systems automatically extract patterns relevant to the NER task from the training set of examples, so they don't require deep language specific knowledge. This makes possible to apply the same NER system to different languages without significant changes in architecture. 

NER task can be considered as a sequence labeling problem. At the moment one of the most common methods to address problems with sequential structure is Recurrent Neural Networks (RNNs) due their ability to store in memory and relate to each other different parts of a sequence.  Thus, RNNs is a natural choice to deal with the NER problem. Up to now, a series of neural models were suggested for NER. To our knowledge on the moment of writing this article the best results for a number of languages such as English, German, Dutch and Spanish  were achieved with a hybrid model combining bi-directional long short-term memory network with conditional random fields (Bi-LSTM + CRF) \cite{Guillaume Lample et al.}. In our study we extended the original work by applying Bi-LSTM + CRF model to NER task in Russian language. We also implemented and experimented with a series of extensions of the NeuroNER model \cite{2017neuroner}. NeuroNER is a different implementation of the same Bi-LSTM + CRF model. However, the realizations of the models might differ in such details as initialization and LSTM cell structure. To reduce training time and improve results, we used the FastText\footnote{An open-source library for learning text representations and text classifiers. URL: https://fasttext.cc/} model trained on Lenta corpus\footnote{A Russian public corpus for some tasks of natural language processing. URL: https://github.com/yutkin/lenta.ru-news-dataset} to obtain external word embeddings. We studied the following models: 
\begin{itemize}
	\item Bi-LSTM (char and word);
	\item Bi-LSTM (char and word) + CRF;
	\item Bi-LSTM (char and word) + CRF + external word embeddings;
    \item Default NeuroNER + char level highway network;
    \item Default NeuroNER + word level highway Bi-LSTM;
    \item Default NeuroNER + char level highway network + word level highway Bi-LSTM.
    
\end{itemize}

To test all models we used three datasets: 
\begin{itemize}
	\item Gareev's dataset \cite{Rinat Gareev et al.};
	\item FactRuEval 2016\footnote{The dataset for NER and Fact Extraction task given at The International Conference on Computational Linguistics and Intellectual Technologies - Moscow 2016};
	\item Persons-1000 \cite{Vlasova at al.}.
\end{itemize}
Our study shows that Bi-LSTM + CRF + external word embeddings model achieves state-of-the-art results for Russian NER task.

%
\section{Neuronal NER Models}
In this section we briefly outline fundamental concepts of recurrent neural networks such as LSTM and Bi-LSTM models. We also describe a hybrid architecture which combines Bi-LSTM with a CRF layer for NER task as well as some extensions of this baseline architecture.
\subsection{Long Short-Term Memory Recurrent Neural Networks}
Recurrent neural networks have been employed to tackle a variety of tasks including natural language processing problems due to its ability to use the previous information from a sequence for calculation of current output. However, it was found \cite{Bengio et al.} that in spite theoretical possibility to learn a long-term dependency in practice RNN models don't perform as expected and suffer from gradient descent issues. For this reason, a special architecture of RNN called Long Short-Term Memory (LSTM) has been developed to deal with the vanishing gradient problem \cite{Sepp Hochreiter et al.}. LSTM replaces hidden units in RNN architecture with units called memory blocks which contain 4 components: input gate, output gate, forget gate and memory cell. Formulas for these components are listed below:
\begin{eqnarray}
i_t = \sigma(W_{ix}x_t + W_{ih}h_{t-1} + b_i),\\
f_t = \sigma(W_{fx}x_t + W_{fh}h_{t-1} + b_f),\\
c_n = g(W_{cx}x_t + W_{ch}h_{t-1} + b_c),\\
c_t = f_t \circ c_{t-1} + i_t \circ c_n,\\
h_t = o_t \circ g(c_t),\\
o_t = \sigma(W_{ox}x_t + W_{oh}h_{t-1} + b_o),
\end{eqnarray}
where $\sigma$, $g$ denote the sigmoid and $tanh$ functions, respectively; $\circ$ is an element-wise product; $W$ terms denotes weight matrices; $b$ are bias vectors; and $i$, $f$, $o$, $c$ denote input gate, forget gate, output gate and cell activation vectors, respectively.
\subsection{Bi-LSTM}
Correct recognition of named entity in a sentence depends on the context of the word. Both preceding and following words matter to predict a tag. Bi-directional recurrent neuronal networks \cite{BiRNN} were designed to encode every element in a sequence taking into account left and right contexts which makes it one of the best choices for NER task. Bi-directional model calculation consists of two steps: (1) the forward layer computes representation of the left context, and (2) the backward layer computes representation of the right context. Outputs of these steps are then concatenated to produce a complete representation of an element of the input sequence. Bi-directional LSTM encoders have been demonstrated to be useful in many NLP tasks such as machine translation, question answering, and especially for NER problem.
\subsection{CRF model for NER task}
Conditional Random Field is a probabilistic model for structured prediction which has been successfully applied in variety of fields, such as computer vision, bioinformatics, natural language processing. CRF can be used independently to solve NER task (\cite{Wenliang Chen et al.}, \cite{Asif Ekbal et al.}). 

The CRF model is trained to predict a vector $\mathbf{y} = \{ y_0, y_1, .., y_T \}$ of tags given a sentence $\mathbf{x} = \{ x_0, x_1, .., x_T \}$. To do this, a conditional probability is computed:
\begin{eqnarray}
p(\mathbf{y}|\mathbf{x}) = \frac{e^{Score(\mathbf{x}, \mathbf{y})}}{\sum_{\mathbf{y}^{'}} e^{Score(\mathbf{x}, \mathbf{y}^{'})}},
\end{eqnarray}
where $Score$ is computed by the formula below \cite{Guillaume Lample et al.}:
\begin{eqnarray}
Score(\mathbf{x}, \mathbf{y}) = \sum_{i=0}^{T} A_{y_i, y_{i+1}} + \sum_{i=1}^{T}P_{i, y_i},
\end{eqnarray}
where $A_{y_i, y_{i+1}}$ denotes the emission probability which represents the score of transition from tag $i$ to tag $j$, $P_{i, j}$ is transition probability which represents the score of the $j^{th}$ tag of the word $i^{th}$.

In the training stage, log probability of correct tag sequence $log(p(\mathbf{y}|\mathbf{x}))$ is maximized.

\subsection{Combined Bi-LSTM and CRF model}
\begin{figure}
	\centering
    \includegraphics[width=0.8\linewidth]{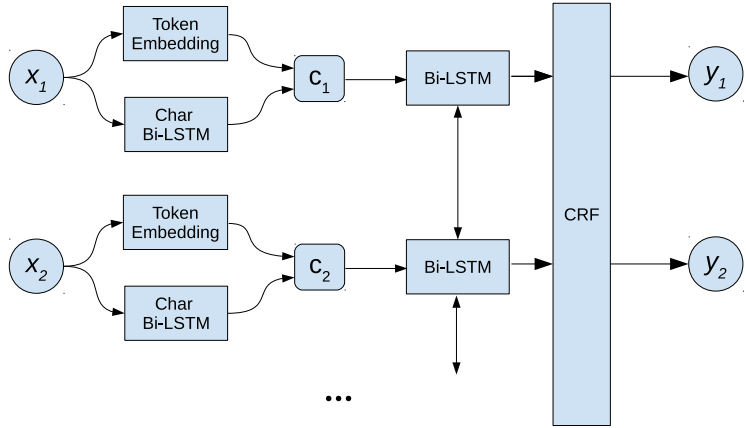} 
	\caption{The architecture of Bi-LSTM neural network for solving NER task. Here, $x_i$ is a representation of word in a sequence. It is fed into character and word level embedding blocks. Then character and word level representations are concatenated into  $c_i$. Bi-LSTM performs conditioning of the concatenated representations on the left and right contexts. Finally CRF layers provide output tag predictions $y_i$.}
	\label{fig:fig1}
\end{figure}

Russian is a morphologically and grammatically rich language. Thus, we expected that a combination of CRF model with a Bi-LSTM neural network encoding \cite{Guillaume Lample et al.} should increase the accuracy of the tagging decisions. The architecture of the model is presented on the figure \ref{fig:fig1} .

In the combined model characters of each word in a sentence are fed into a Bi-LSTM network in order to capture character-level features of words. Then these character-level vector representations are concatenated with word embedding vectors and fed into another Bi-LSTM network. This network calculates a sequence of scores that represent likelihoods of tags for each word in the sentence. To improve accuracy of the prediction a CRF layer is trained to enforce constraints dependent on the order of tags. For example, in the IOB scheme (I -- Inside, O -- Other, B -- Begin) tag I never appears at the beginning of a sentence, or “O I B O” is an invalid sequence of tags.


Full set of parameters for this model consists of parameters of Bi-LSTM layers (weight matrices, biases, word embedding matrix) and transition matrix of CRF layer. All these parameters are tuned during training stage by back propagation algorithm with stochastic gradient descent. Dropout is applied to avoid over-fitting and improve the system performance. 

\subsection{Neuro NER Extensions}
NeuroNER is an open-source software package for solving NER tasks. The neural network architecture of NeuroNER is similar to the architecture proposed in the previous section.

\begin{figure}
	\centering
    \includegraphics[width=0.7\linewidth]{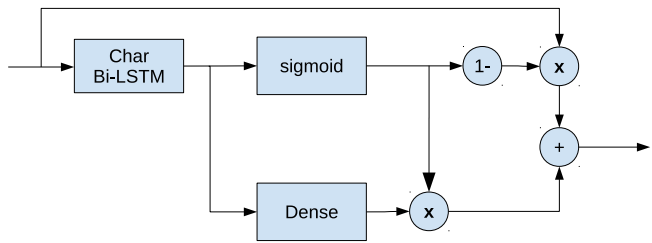}
	\caption{Highway network on top of the character embedding network. The Dense layer serves for compute higher level representations of the input. The sigmoid layer computes gate values. This values are used to dynamically balance between high and low level representations. Block (1-) subtracts input from 1, and block (x) perform multiplication of the inputs.} 
	\label{fig:fig2}
\end{figure}

Inspired by success of character aware networks approach \cite{Kim et al.} we extended NeuroNER model with a highway layer on top of the Bi-LSTM character embedding layer. This extension is depicted on figure \ref{fig:fig2}. Dense layer makes character embedding network deeper. The carry gate presented by sigmoid layer provides a possibility to choose between dense and shortcut connections dynamically. A highway network can be described by the following equation:

\begin{equation}
\mathbf{y} = H(\mathbf{x}, \mathbf{W_H}) \cdot G(\mathbf{x}, \mathbf{W_G}) + \mathbf{x} \cdot (1 - G(\mathbf{x}, \mathbf{W_G}))
\end{equation}
where \(\mathbf{x} \) is the input of the network, \(H(\mathbf{x}, \mathbf{W_H})  \) is the processing function, \( G(\mathbf{x}, \mathbf{W_G}) \) is the gating function. The dimensionality of \(\mathbf{x} \), \(\mathbf{y} \), \( G(\mathbf{x}, \mathbf{W_G}) \), and  \(H(\mathbf{x}, \mathbf{W_H})  \) must be the same.

\begin{figure}
	\centering
	\includegraphics[width=0.6\linewidth]{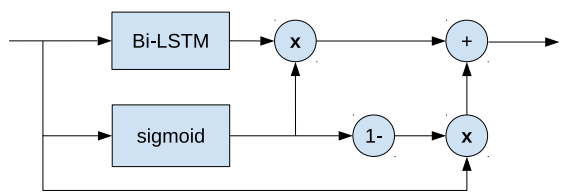}
	\caption{Highway LSTM network. Here sigmoid gate layer is used to dynamically balance between input and output of the Bi-LSTM layers. The gating applied to the each direction separately.}
	\label{fig:fig3}
\end{figure}

Another extension of NeuroNER we implemented is a Bi-LSTM highway network \cite{Pundak et al.}. The architecture of this network is quite similar to the character-aware highway network. However, the carry gate is conditioned on the input of the LSTM cell. The gate provides an ability to dynamically balance between raw embeddings and context dependent LSTM representation of the input. The scheme of our implementation of the highway LSTM is depicted in figure \ref{fig:fig3}.

\section{Experiments}
\subsection{Datasets} \label{datasets}

Currently, there are a few Russian datasets created for the purpose of developing and testing NER systems. We trained and evaluated models on the three Russian datasets:

\begin{itemize}
	
\item Dataset received from Gareev et al. \cite{Rinat Gareev et al.} contains 97 documents collected from ten top cited "Business" feeds in Yandex "News" web directory. IOB tagging scheme is used in this data sets, and entity types are Person, Organization, Other.

\item The FactRuEval 2016 corpus \cite{Starostin A. et al.} contains news and analytical texts in Russian. Sources of the dataset are Private Correspondent\footnote{http://www.chaskor.ru/} web site and Wikinews\footnote{http://ru.wikinews.org}. Topics of the texts are social and political. Tagging scheme is IOB.

\item Person-1000 \cite{Vlasova at al.} is a Russian news corpus with marked up person named entities. This corpus contains materials from the Russian on line news services.

\end{itemize}

Statistics on these datasets are provided in the table \ref{table:corpuses}.

\begin{table}
\caption{Statistics on datasets}
\label{table:corpuses}
\begin{center}
\begin{tabular}{|*{6}{c|}}
 \hline
Datasets & Tokens & Words and numbers & Persons & Organizations & Locations  \\
 \hline
 FactRuEval 2016 & 90322 & 73807 & 2087 & 1181 & 2686 \\
 \hline
 Gareev's dataset   & 44326  &35116   & 486 & 1317  & - \\
 \hline
 Persons-1000 & 284221 & 224446 & 10600 & - & -         \\
 \hline
\end{tabular}
\end{center}
\end{table}

\subsection{External Word Embedding}
$News$ and $Lenta$ are two external word embeddings we used to initialize lookup table for the training step.

$News$\footnote{Word embeddings, which are available to  download from http://rusvectores.org} is a Russian word embeddings introduced by Kutuzov, et al. \cite{Kutuzov A. et al.}. Corpus for this word embedding is a set of Russian news (from September 2013 until November 2016). Here are more details about $news$:
\begin{itemize}
\item Corpus size: near 5 billion words
\item Vocabulary size: 194058
\item Frequency threshold: 200
\item Algorithm: Continuous Bag of Words
\item Vector size: 300 
\end{itemize}

$Lenta$ is a publicly available corpus of unannotated Russian news. This corpus consists of 635000 news from Russian online news resource lenta.ru. The size of the corpus is around 46 million words. The corpus spans vocabulary of size 376000 words.

To train embeddings on this corpus, we use skip-gram algorithm enriched with subword information  \cite{Bojanowski at al.}. Parameters of the algorithm were the following:

\begin{itemize}
\item Vector size: 100
\item Minimal length of char n-gram: 3
\item Maximal length of char n-gram: 6
\item Frequency threshold: 10
\end{itemize}

\subsection{Results} \label{results}
The purpose of the first experiment was to compare tagging accuracy of three implementations: Bi-LSTM, Bi-LSTM + CRF, Bi-LSTM + CRF + external word embedding $news$. To do this, we evaluated these implementations on the $Gareev's \, dataset$. Parameters of the dataset and hyper-parameters of the models are listed below:
\begin{itemize}
\item Word embedding dimension: 100
\item Char embedding dimension: 25
\item Dimension of hidden layer: 100 (for each LSTM: forward layer and backward layer)
\item Learning method: SGD, learning rate: 0.005
\item Dropout: 0.5
\item Number of sentences: 2136 (for training/ validation/ testing: 1282/ 427/ 427)
\item Number of words: 25372 (unique words: 7876). 7208 words (account for 91.52\% of unique words) was initialized with pre-trained embedding $Lenta$
\item epochs: 100
\end{itemize}
We used ConllEval\footnote{A Perl script was used to evaluate result of processing CoNLL-2000 shared task: http://www.cnts.ua.ac.be/conll2000/chunking/conlleval.txt} to calculate metrics of performance. 

The result is shown in the Table \ref{table: ThreeModelsOnGareevDataset}. One can see that adding CRF layer significantly improved prediction. Besides that, using external word embeddings also reduced training time and increased tagging accuracy. Due to absence of lemmatization in our text processing pipeline $news$ embeddings matched only about 15\% of words in the corpus, embeddings for other words were just initialized randomly. Therefore, the improvement was not really significant and prediction for $Organization$ type was even lower with $news$ embeddings. To deal with this problem, in the second experiment we decided to use FastText trained on $Lenta$ corpus in order to build an external word embedding. After that, we used this embedding to train on Gareev's dataset one more time using the same configuration with the previous experiment.


\begin{table}
\caption{Tagging results on Gareev's dataset}
\label{table: ThreeModelsOnGareevDataset}
\begin{center}
\begin{tabular}{|*{10}{c|}}
\hline
\multicolumn{1}{|c}{\multirow{2}{*}{Model}} & \multicolumn{3}{|c}{Person} & \multicolumn{3}{|c}{Organization} & \multicolumn{3}{|c|}{Overall}\\
\cline{2-10}
 & P & R & F & P & R & F & P & R & F \\ 
 \hline
Bi-LSTM & 67.11 & 78.46 & 72.34 & 76.56 & 72.59 & 74.52 & 73.04 & 74.50 & 73.76 \\
\hline 
Bi-LSTM CRF & 92.93 & 86.79 & 89.76 & 85.24 & \textbf{81.91} & 83.54 & 87.30 & 83.25 & 85.22 \\

\hline 
\makecell{Bi-LSTM CRF + \\ $news$ word emb.} & 95.05 & 90.57 & 92.75 & 85.13 & 81.21 & 83.12 & 87.84 & 83.76 & 85.75 \\ 
\hline
\makecell{Bi-LSTM CRF + \\ $Lenta$ word emb.} & \textbf{95.60} & \textbf{94.57} & \textbf{95.08} & \textbf{87.40} & 81.62 & \textbf{84.41} & \textbf{89.57} & \textbf{84.89} & \textbf{87.17} \\

\hline
\end{tabular}
\end{center}
\end{table}

Table \ref{table:ConfusionMatrixOnGareevDataset} shows the confusion matrix on the test set. We also experimented on two other datasets: Persons-1000, FactRuEval 2016. The summary of experiments on these datasets are shown in the Table \ref{table:TaggingResultsOn3DatasetsUsingLenta}.

\begin{table}
\caption{The confusion matrix on the test set of Gareev's dataset}
\label{table:ConfusionMatrixOnGareevDataset}
\begin{center}
\begin{tabular}{|*{8}{c|}}
 \hline
 Named Entity & Total & O & I-ORG & B-ORG & B-PER & I-PER & Percent \\
 \hline
 O   & 7688   & 7647    & 19    & 22     & 0     & 0  & 99.467 \\
 \hline
 I-ORG   & 308    & 36   & 268    & 3    & 1     & 0   	& 87.013 \\
 \hline
 B-ORG 	& 272    & 38    & 2   & 229    & 2     & 1   	& 84.191 \\
 \hline
 B-PER  & 92     & 3     & 1     & 0    & 88    & 0  & 95.652 \\
 \hline
 I-PER   & 69    & 2     & 5    & 0     & 0    & 62   & 89.855 \\
 \hline
\end{tabular}
\end{center}
\end{table}

\begin{table}
\caption{Tagging results of Bi-LSTM + CRF + $Lenta$ word embedding on three datasets: Gareev's dataset, FactRuEval 2016, Persons-1000}
\label{table:TaggingResultsOn3DatasetsUsingLenta}
\begin{center}
\begin{tabular}{|*{7}{c|}}
\hline
\multicolumn{1}{|c}{\multirow{2}{*}{Datasets}} & \multicolumn{3}{|c}{Validation set} & \multicolumn{3}{|c|}{Test set} \\
\cline{2-7}
& P & R & F & P & R & F \\
\hline
FactRuEval 2016 & 84.39 & 81.11 & 82.72 & 83.88 & 80.40 & 82.10 \\
\hline
Gareev's dataset & 90.99 & 86.94 & 88.92 & 89.57 & 84.89 & 87.17 \\
\hline
Persons-1000 & 98.97 & 98.20 & 98.58 & 99.43 & 99.09 & 99.26 \\
\hline
\end{tabular}
\end{center}
\end{table}

We compare Bi-LSTM + CRF + $Lenta$ model and other published results as well as NeuroNER and its extensions on three datasets mentioned in the subsection \ref{datasets}. Results are presented in the table \ref{table:comparison}.  Bi-LSTM + CRF + $Lenta$ model significantly outperforms other approaches on Gareev's dataset and Persons-1000. However, the result on FactRuEval 2016 dataset is not as high as we expected.

\begin{table}
\caption{Performance of different models across datasets}
\label{table:comparison}
\begin{center}
\begin{tabular}{|*{10}{c|}}
\hline
\multicolumn{1}{|c}{\multirow{2}{*}{Models}} & \multicolumn{3}{|c}{Gareev's dataset} & \multicolumn{3}{|c}{Persons-1000} & \multicolumn{3}{|c|}{FactRuEval 2016} \\
\cline{2-10}
& P & R & F & P & R & F & P & R & F \\
\hline
Gareev et al. \cite{Rinat Gareev et al.} & 84.10 & 67.98 & 75.05  & - & - & - & - & - & - \\
\hline
Malykh et al. \cite{Valentin Malykh et al.} & 59.65 & 65.70 & 62.49 & - & - & - & - & - & - \\
\hline
Trofimov \cite{Trofimov}  & - & - & - & 97.26 & 93.92 & 95.57  & - & - & - \\
\hline
Rubaylo et al. \cite{Rubaylo et al.}  & - & - & - & - & - & - & 77.70 & 78.50 & 78.13 \\
\hline

Sysoev et al.\cite{Sysoev et al.} & - & - & - & - & - & - & \textbf{88.19} & 64.75 & 74.67 \\
\hline

Ivanitsky et al. \cite{Ivanitskiy et al.}  & - & - & - & - & - & - & - & - & \textbf{87.88} \\
\hline
Mozharova et al. \cite{Mozharova et al.}  & - & - & - & - & - & 97.21  & - & - & - \\
\hline
\hline
NeuroNER                & 88.19 & 82.73 & 85.37 & 96.38  & 96.83 & 96.60 & 80.49  & 79.23 & 79.86 \\ 

\hline
\makecell{NeuroNER + \\ Highway char}  & 85.75 & \textbf{88.40} & 87.06 & 96.56 & 97.11 & 96.83 & 80.59 & 80.72 & 80.66 \\
\hline
\makecell{NeuroNER + \\ Highway LSTM}  & 84.35 & 81.96 & 83.14 & 96.49 & 97.19 & 96.84 & 81.09 & 79.31 & 80.19 \\
\hline
\makecell{NeuroNER + \\ Highway char +\\ Highway LSTM}   & 83.33 & 85.05 & 84.18 & 96.74 & 96.83 & 96.78 & 79.13 & 78.76 & 78.95 \\
\hline
\hline
\makecell{Bi-LSTM + CRF + \\$Lenta$} & \textbf{89.57} & 84.89 & \textbf{87.17} & \textbf{99.43} & \textbf{99.09} & \textbf{99.26} & 83.88 & \textbf{80.84} & 82.10 \\
\hline
\end{tabular}
\end{center}
\end{table}

\section{Discussion}
Traditional approaches to Russian NER heavily relied on hand-crafted rules and external resources. Thus regular expressions and dictionaries were used in \cite{Trofimov} to solve the task. The next step was application of statistical learning methods such as conditional random fields (CRF) and support vector machines (SVM) for entity classification. CRF on top of linguistic features considered as a baseline in the study of \cite{Rinat Gareev et al.}. Mozharova and Loukachevitch \cite{Mozharova et al.} proposed two-stage CRF algorithm. Here, an input for the CRF of the first stage was a set of hand-crafted linguistic features. Then on the second stage the same input features were combined with a global statistics calculated on the first stage and fed into CRF. Ivanitskiy et al. \cite{Ivanitskiy et al.} applied SVM classifier to the distributed representations of words and phrases. These representations were obtained by extensive unsupervised pre-training on different news corpora. Simultaneous use of dictionary based features and distributed word representations was presented in \cite{Sysoev et al.}. Dictionary features were retrieved from Wikidata and word representations were pre-trained on Wikipedia. Then these features were used for classification with SVM.

At the moment deep learning methods are seen as the most promising choice for NER. Malykh and Ozerin \cite{Valentin Malykh et al.} proposed character aware deep LSTM network for solving Russian NER task. A distinctive feature of this work is coupling of language modeling task with named entity classification.

In our study we applied current state of the art neural network based model for English NER to known Russian NER datasets. The model consists of three main components such as bi-directional LSTM, CRF and external word embeddings. Our experiments demonstrated that Bi-LSTM alone was slightly worse than CRF based model of \cite{Rinat Gareev et al.}. Addition of CRF as a next processing step on top of Bi-LSTM layer significantly improves model's performance and allow to outperform the model presented in \cite{Rinat Gareev et al.}. The difference of Bi-LSTM + CRF model from the model presented in \cite{Rinat Gareev et al.} is trainable feature representations. Combined training of Bi-LSTM network on the levels of words and characters gave better results then manual feature engineering in \cite{Rinat Gareev et al.}. 

Distributed word representations are becoming a standard tool in the field of natural language processing. Such representations are able to capture semantic features of words and significantly improve results for different tasks. When we encoded words with $news$ or $Lenta$ embeddings results were consistently better for all three datasets. Up to now, the prediction accuracy of the Bi-LSTM + CRF + $Lenta$ model outperforms published models on Gareev's dataset and Persons-1000. However, the results of both  Bi-LSTM + CRF + $Lenta$ and NeuroNER models on the FactRuEval dataset were better then results reported in \cite{Sysoev et al.} and \cite{Rubaylo et al.} but not as good as SVM based model reported in \cite{Ivanitskiy et al.}. 

In spite the fact that both models we tested have the same structure, performance of NeuroNER \cite{2017neuroner}  is a bit lower than Bi-LSTM+CRF model \cite{Guillaume Lample et al.}. This issue can be explained by different strategies for initialization of parameters.

Our extension of the baseline model with a highway network for character embedding provides moderate performance growth in nearly all cases. Implementation of the Bi-LSTM highway network for tokens resulted in a slight increase of performance for Persons-100 and FactRuEval 2016 datasets and a decrease of performance for Gareev's dataset. Simultaneous extension of the NeuroNER with character and token Bi-LSTM highway networks results in the drop of performance in the most of the cases.

We think that results of LSTM highway network can be improved by different bias initialization and deeper architectures. In the current work the highway gate bias was initialized with 0 vector. However, bias could be initialized to some negative value. This initialization will force the network to prefer processed path to the raw path. Furthermore, stacking highway LSTM layers might improve results allowing a network dynamically adjust complexity of the processing. Alternatively, character embedding network can be built using convolutional neural networks (CNN) instead of LSTM. A number of authors \cite{Kim et al.,Tran et al.} reported promising results with a character level CNN. Another promising extension of presented architecture is an attention mechanism \cite{BengioAttn}. For NER task this mechanism can be used to selectively attend to the different parts of the context for each word giving additional information for the tagging decision.

\section{Conclusions}
Named Entity Recognition is an important stage in information extraction tasks. Today, neural network methods for solving NER task in English demonstrate the highest potential. For Russian language there are still a few papers describing application of neural networks to NER. We studied a series of neural models starting from vanilla bi-directional LSTM then supplementing it with conditional random fields, highway networks and finally adding external word embeddings. For the first time in the literature evaluation of models were performed across three Russian NER datasets.  Our results demonstrated that (1) basic Bi-LSTM model is not sufficient to outperform existing state of the art NER solutions, (2) addition of CRF layer to the Bi-LSTM model significantly increases it's quality, (3) pre-processing the word level input of the model with external word embeddings allowed to  improve performance further and achieve state-of-the-art for the Russian NER.

\section{Acknowledgments}
The statement of author contributions. AL conducted initial literature review, selected a baseline (Bi-LSTM + CRF) model, prepared datasets and run experiments under supervision of MB.  AM implemented and studied extensions of the NeuroNER model. AL drafted the first version of the paper. AM added a review of works related to the Russian NER and materials related to the NeuroNER modifications. MB, AL and AM edited and extended the manuscript.

This work was supported by National Technology Initiative and PAO Sberbank project ID 0000000007417F630002.

%
%

\end{document}